\pdfoutput=1

\documentclass[11pt]{article}

\usepackage{amsmath,amsfonts,bm}

\def\eqref#1{equation~\ref{#1}}

\def\1{\bm{1}}

\DeclareMathAlphabet{\mathsfit}{\encodingdefault}{\sfdefault}{m}{sl}
\SetMathAlphabet{\mathsfit}{bold}{\encodingdefault}{\sfdefault}{bx}{n}

\DeclareMathOperator*{\argmax}{arg\,max}
\DeclareMathOperator*{\argmin}{arg\,min}

\usepackage[]{acl}

\usepackage{times}
\usepackage{latexsym}

\usepackage[utf8]{inputenc} 
\usepackage[T1]{fontenc}    
\usepackage{hyperref}       
\usepackage{url}            
\usepackage{booktabs}       
\usepackage{amsfonts}       
\usepackage{nicefrac}       
\usepackage{microtype}      
\usepackage{xcolor}         
\usepackage{bm,bm,bbm,dsfont}
\usepackage{comment}
\usepackage{pgfplots}
\usepackage{arydshln}
\usepackage{pifont}
\usepackage{amsfonts}
\usepackage{multicol,multirow}
\usepackage{amsmath, amssymb, amsthm}
\usepackage[ruled]{algorithm2e}

\usepackage{bm,bbm,dsfont}
\pgfplotsset{width=7cm,compat=1.13}

\usepackage[T1]{fontenc}
\usepackage[utf8]{inputenc}

\usepackage{microtype}
\newcommand{\sts}{{{\textsc{Seq2Seq}}}\xspace}

\usepackage{microtype}
\definecolor{ao}{rgb}{0.0, 0.5, 0.0}
\definecolor{asparagus}{rgb}{0.53, 0.66, 0.42}
\definecolor{amber}{rgb}{1.0, 0.49, 0.0}
\definecolor{alizarin}{rgb}{0.82, 0.1, 0.26}
\definecolor{applegreen}{rgb}{0.55, 0.71, 0.0}
\definecolor{amethyst}{rgb}{0.6, 0.4, 0.8}
\definecolor{auburn}{rgb}{0.43, 0.21, 0.1}

\title{Paraphrase Generation as Unsupervised Machine Translation}
\date{}
\author{
Xiaofei Sun$^{1,2}$, Yufei Tian$^{3}$, Yuxian Meng$^{2}$, Nanyun Peng$^{3}$\\
{\bf Fei Wu$^{1,5}$, Jiwei Li$^{1,2}$ and Chun Fan$^{4}$}\\
$^1$Zhejiang University, $^2$Shannon.AI, $^3$University of California, Los Angeles \\
$^4$Peng Cheng Laboratory,
$^4$National Biomedical Imaging Center, Peking University \\
$^4$ Computer Center, Peking University, $^5$Shanghai AI Laboratory  \\ $^5$Shanghai Institute for Advanced Study of Zhejiang University\\
xiaofei\_sun@zju.edu.cn, yufeit@ucla.edu, jiwei\_li@shannonai.com
}

\begin{document}
\maketitle
\begin{abstract}
In this paper, we propose a new paradigm for paraphrase generation by treating the task as unsupervised machine translation (UMT) based on the assumption that there must be pairs of sentences expressing the same meaning in a large-scale unlabeled monolingual corpus. The proposed paradigm first splits a large unlabeled corpus into multiple clusters, and trains  multiple UMT models using  pairs of these clusters. Then based on the  paraphrase pairs produced by these UMT models, a unified surrogate model can be trained to serve as the final \sts model to generate paraphrases, which can be directly used for test in the unsupervised setup, or be finetuned on labeled datasets in the supervised setup. The proposed method offers merits over machine-translation-based paraphrase generation methods, as it avoids  reliance on bilingual sentence pairs. It also allows human intervene with the model so that more diverse paraphrases can be generated using different filtering criteria.
Extensive experiments on existing paraphrase dataset for both the supervised and unsupervised setups demonstrate the effectiveness the proposed paradigm. 
\footnote{Accepted by COLING 2022.}
\end{abstract}

\section{Introduction}
The goal of paraphrase generation \cite{prakash2016neural,cao2016joint,ma-etal-2018-query,wang2018task} is to generate a sentence semantically identical to a given input sentence but with variations in lexicon or syntax.  
It has been applied to various downstream NLP tasks such as parsing \citep{berant2014semantic}, question answering \citep{dong2017learning}, summarization \citep{barzilay2004information} and machine translation \citep{callison2006improved}.

Building a strong paraphrase generation system usually requires massive amounts of high-quality annotated paraphrase pairs, but existing labeled datasets \citep{lin2014microsoft,fader-etal-2013-paraphrase,lan-etal-2017-continuously} are either of small sizes or restricted in narrow domains.
To avoid such a heavy reliance on labeled datasets, recent works have explored unsupervised methods \citep{li-etal-2018-paraphrase,fu2019paraphrase,Siddique_2020} to generate paraphrase without annotated training data, among which the back-translation based model is an archetype \citep{mallinson2017paraphrasing,sokolov2020neural}. 
It borrows the idea of back-translation (BT) in machine translation \citep{sennrich2016back-translation} where the model first translates a sentence $s_1$ into another sentence $s_2$ in a different language  (e.g., En$\rightarrow$Fr), and then translates $s_2$ back to $s_1$. In this way, the model is able to generate paraphrases by harnessing 
 bilingual datasets, removing the need for label paraphrase data. 
 
However, BT-based models for paraphrase generation have the following severe issues: 
firstly, BT-based systems  rely on external resources, i.e., 
 bilingual datasets, making them hard to be  applied to 
  languages whose bilingual datasets are hard to obtain.
Secondly,
translation errors, such as duplicate words \citep{Holtzman2020TheCC}, missing words \citep{luong2015addressing} and polysemous words \citep{rios-gonzales-etal-2017-improving}, will accumulate during the forward and backward translations, resulting in inferior performances.
Thirdly, machine translation models work like blackboxs, making it hard for humans to intervene with the model and control the generation process.

In this work, we propose to address these problems based on the assumption that there must be pairs of sentences expressing the same meaning in a large-scale unlabeled corpus.
Inspired by  unsupervised machine translation (UMT) models, which
 align semantic spaces of two languages using monolingual data, we propose a pipeline system to generate paraphrases following two stages:
(1) splitting a large-scale monolingual corpus into multiple clusters/sub-datasets, on which UMT models are trained based on pairs of these sub-datasets; 
and (2) training a unified surrogate  model based on the paraphrase pairs produced by the trained multiple UMT models, where  we can design filtering functions to remove the pairs with undesired properties. The unified surrogate model can be then directly used for paraphrase generation in the unsupervised setup, or be finetuned on labeled datasets in the supervised setup.

The proposed framework provides the following merits over existing BT-based methods: 
(1) it is purely based on a large-scale monolingual corpus, which removes the  reliance on bilingual datasets;
(2) the trained unified model is able to generate paraphrases end-to-end, which avoids the issue of error accumulation that exists in vanilla BT-based models;
and (3) human interventions can take place in the filtering step, which gives finer-grained controls over the generated paraphrases.

We conduct extensive experiments on a wide range of paraphrase datasets to evaluate the effectiveness of the proposed framework, and we are able to observe  performance boosts against strong baselines in both  supervised and unsupervised setups.

\section{Related Work}  
\paragraph{Paraphrase Generation}
Methods for paraphrase generation usually fall into two categories: supervised and unsupervised approaches.
Supervised methods for paraphrase generation rely on annotated paraphrase pairs.
\citet{xu2018d,qian2019exploring} employed distinct semantic style embeddings to generate diverse paraphrases, and 
\citet{iyyer2018adversarial,li-etal-2019-decomposable,chen2019controllable,goyal2020neural} proposed to use different syntactic structure templates.
A line of work \citep {mallinson2017paraphrasing,sokolov2020neural} formalized paraphrase generation as machine translation.
Unsupervised paraphrase generation is primarily based on reinforcement learning (RL)  generative models \cite{ranzato2015sequence,li2016deep}.
RL optimizes certain criteria, e.g. BLEU, to reward paraphrases with higher quality \citep{li-etal-2018-paraphrase,Siddique_2020}. 
\citet{bowman-etal-2016-generating,yang2019end} trained a variational auto-encoder (VAE) \citep{kingma2013auto} to generate paraphrases. 
Other unsupervised methods for paraphrase generation include VAE (VQ-VAE) \citep{roy2019unsupervised}, latent bag-of-words alignment \citep{fu2019paraphrase} and simulated annealing \citep{liu2019unsupervised}. Adapting large-scale pretraining \cite{devlin2018bert,radford2018improving,liu2019roberta,clark2020electra,sun2021chinesebert}
to paraphrase generation has been recently investigated \citep{witteveen2019paraphrasing,hegde2020unsupervised,niu2020unsupervised,meng2021conrpg} and has shown promising potentials to improve generation quality.
Our work is  distantly related to unsupervised text style transfer \citep{hu2017toward,mueller2017sequence,shen2017style,li2018delete,fu2018style}, where the model alters a specific text attribute of an input sentence  (such as sentiment) while preserving other text attributes.

Regarding soliciting large-scale paraphrase datasets,
\newcite{bannard2005paraphrasing}
used statistical machine translation
methods  
obtain  paraphrases in parallel
text, the technique of which is scaled up by \newcite{ganitkevitch2013ppdb}
to
produce the Paraphrase Database
(PPDB). 
\newcite{wieting2017learning} 
 translate the non-English side of parallel text to obtain paraphrase pairs. 
\newcite{wieting2017paranmt} collected paraphrase dataset with million of pairs via machine translation. 
\newcite{hu2019improved,hu2019parabank} 
produced paraphrases from a bilingual corpus based on the techniques of negative constraints and inference sampling. 


\paragraph{Unsupervised Machine Translation}
Unsupervised Machine Translation(UMT) has been an active research direction in NLP 
  \citep{ravi-knight-2011-deciphering}.
Pioneering work for unsupervised neural machine translation used denoising auto-encoders and back-translation \citep{sennrich2016back-translation} to iteratively refine the generated translation.
\citet{artetxe2017unsupervised} proposed to use a shared encoder to encode source input sentences from different languages.
\citet{lample2017unsupervised} additionally used adversarial and cross-domain training objectives to better identify different language domains.
\citet{yang-etal-2018-unsupervised} relaxed the strategy of sharing the entire encoder in \citet{artetxe2017unsupervised} by building independent encoders to maintain unique characteristics of each language.
Another line of work for UMT is to combine statistical machine translation (SMT) and NMT.
\citet{artetxe2018unsupervised,lample2018phrase} built a phrase-level mapping table from the source language to the target language. 
Following works improved UMT by combining SMT and NMT in different ways, such as warming up an NMT model with a trained SMT model \citep{marie2018unsupervised,artetxe2019effective} and using SMT as posterior regularization \citep{ren2019unsupervised}. Other works involve initializing the model using retrieved semantically similar sentence pairs  \citep{wu2019extract,ren-etal-2020-retrieve,sun2021sentence}, using auxiliary parallel data \citep{li2020reference,garcia2020multilingual} and pretraining on large-scale multi-lingual data \citep{lample2019cross,song2019mass,liu2020multilingual,zhu2020incorporating}.


\section{Background for Unsupervised Machine Translation}
We use the unsupervised machine translation (UMT) framework proposed by \citet{lample2017unsupervised} as the backbone. We  briefly go though the model structure in this section.
Let $C_{src}$ and $C_{tgt}$ respectively denote the monolingual dataset for the source and target language, on which a translation model $M$ is trained to 
to generate target sequences $y$ based on source sequences $x$, $y=M(x)$. 
The model is first initialized by training in a word-by-word translation manner using a parallel dictionary. 
The initial parallel dictionary is thus a word being translated to itself. 
Next, the model is iteratively trained based on a denoising auto-encoding (DAE), back-training (BT) and adversarial learning (AL). 
DAE allows the model to reconstruct the translation from a noisy input sentence by dropping and swapping words in the original sentence. 
The training objective of DAE is given by:
\begin{equation}
  \begin{aligned}
    \mathcal{L}^l_\text{DAE}=\mathbb{E}_{x\sim C_l,\hat{x}\sim d(e(N(x),l),l)}[\Delta(\hat{x},x)]
  \end{aligned}
\end{equation}
where $l=src$ or $l=tgt$ specifies the language, $N(x)$ is a noisy version of $x$, $e$ and $d$ respectively means encoding and decoding, and $\Delta$ measures the difference between the two sequences, which is the cross-entropy loss in this case. 
BT encourages the model to reconstruct the input sentence $x$ from $N(y)$, a corrupted version of the model's translation $y=M(x)$. The training objective is given by:
\begin{equation}
  \begin{aligned}
    \mathcal{L}^{l_1\to l_2}_\text{BT}=\mathbb{E}_{x\sim C_{l_1},\hat{x}\sim d(e(N(M(x)),l_2),l_1)}[\Delta(\hat{x},x)]
  \end{aligned}
\end{equation}
AL uses a discriminator to distinguish the language from the encoded latent representations,
and by doing so, the model is able to better map two languages into the same latent space.
The discriminative training objective is given by:
\begin{equation}
  \begin{aligned}
    \mathcal{L}^l_\text{Dis}=-\mathbb{E}_{(x,l)}[\log p(l|e(x,l))]
  \end{aligned}
\end{equation}
The encoder is trained to fool the discriminator so that the encoder and the discriminator perform together in an adversarial style \citep{goodfellow2014generative}:
\begin{equation}
  \begin{aligned}
    \mathcal{L}^{l_1\to l_2}_\text{Adv}=-\mathbb{E}_{(x_1,l_1)}[\log p(l_2|e(x_1,l_1))]
  \end{aligned}
\end{equation}
The final training objective is given by:
\begin{equation}
  \begin{aligned}
    &\mathcal{L}=\lambda_1[\mathcal{L}^{l_1}_\text{DAE}+\mathcal{L}^{l_2}_\text{DAE}]+\lambda_2[\mathcal{L}^{l_1\to l_2}_\text{BT}+\mathcal{L}^{l_2\to l_1}_\text{BT}]\\&+\lambda_3 [\mathcal{L}^{l_1\to l_2}_\text{Adv}+\mathcal{L}^{l_2\to l_1}_\text{Adv}]
  \end{aligned}
\end{equation}
The discriminative loss $\mathcal{L}^l_\text{Dis}$ is alternatively optimized with $\mathcal{L}$ to train the discriminator.
We follow \citet{lample2017unsupervised} to implement each of the UMT models.
We used the transformer-large \cite{vaswani2017attention} as the backbone instead of LSTMs in \newcite{bahdanau2014neural}.

\section{Model}
The core idea of the proposed strategy is to use two subdatasets from a large monolingual corpus $C$
and train unsupervised NMT models based on the two subdatasets.
The path towards this goal naturally constitutes two modules: (1) constructing two subdatasets  $C_{src}$ and $C_{tgt}$ from
 $C$; and (2)  training the UMT model based on $C_{src}$ and $C_{tgt}$.

\subsection{Dataset Split}
The crucial part in the framework is how to build the two subdatasets, on which the unsupervised NMT model is trained.
 To this end, we propose to (1) first construct candidates $\{c_1, c_2, ..., c_K\}$ for $C_{src}$ and $C_{tgt}$ from $C$ based on the  clustering models; and (2) selecting  $C_{src}$ and $C_{tgt}$. Based on $C_{src}$ and $C_{tgt}$,  UMT models will be trained. 
We use two criteria for clustering, LDA \citep{blei2003latent} 
and K-means clustering. The number of clusters/topics $K$ is set to 80.\footnote{Here we use ``topic" and ``cluster" interchangeably.} 

\paragraph{LDA Clustering}
For LDA, we use Gibbs sampling and iterate over the entire corpus 5 times in total. In the last round,  a sentence is assigned to the topic/cluster which has the largest probability of generating it. 
In LDA, each cluster is characterized as a distribution over the vocabulary. The distance between two subset $c_m$, $c_n$ is the Jensen–Shannon (JS) divergence between the two distributions over the vocabulary: 
\begin{equation}
\begin{aligned}
\text{Dis}(c_m, c_n) &= \text{KL}(c_m||c_n) + \text{KL}(c_n||c_m)  \\
\text{KL}(c_m||c_n) & = -\sum_{v\in V} p(v|c_m) \log\frac{ p(v|c_m) }{ p(v|c_n) }\\
\text{KL}(c_n||c_m) & = -\sum_{v\in V} p(v|c_n) \log\frac{ p(v|c_n) }{ p(v|c_m) }
\end{aligned}
\end{equation}
Since topics clustered by LDA can be incoherent (e.g., the clustering of stop words), we ask humans to examine the top words of the topics, and discard meaningless clusters.  

\paragraph{K-means Clustering}
For K-means, we use the average of the top layer embeddings from BERT \citep{devlin2018bert} to represent the sentence. 
Let $h_s$ denote the sentence representation for the sentence $s$. 
We run the hard K-means model on the corpus,
where the distance between a  sentence and the cluster center is the $L_2$ distance between the two vector representations. 

The LDA and K-means methods described above focus more on 
the situation that
 centers of two clusters are far away, but not individual sentences belonging to different clusters are different.
These two focuses are correlated, but not exactly the same.
The JS divergence for LDA clusters and $L_2$ distance for K-means clusters will be updated after the post-processing stage. 
LDA and K-means algorithms are performed 
on part of the 
the CommonCrawl corpus containing 10 billion English tokens. 

\subsection{UMT Training on $C_{src}$ and $C_{tgt}$}    
  
 \subsubsection{Multiple UMT Models} 
  We can randomly pick 
one pair of subsets 
 from  $\{c_1, ..., c_K\}$ as $C_{src}$ and $C_{tgt}$, 
  on which  a single UMT model will be trained. 
The problem with single UMT model is obvious: each subset in 
$\{c_1, c_2, ..., c_K\}$ potentially represents a specific domain. 
The UMT model trained on the single $C_{src}$ can thus only be able to properly paraphrase sentences from the 
   $C_{src}$ domain. 
To cover the full domain, we propose to train $K$ UMT models, denoted by $\{M_1, M_2, ..., M_K\}$, 
where $K$ is the  number of clusters. 
Each of the trained UMT models uses a different $c\in \{c_1, c_2, ..., c_K\}$ as $C_{src}$, paired with a randomly selected $C_{tgt}$.

\begin{figure*}
  \centering
  \includegraphics[scale=0.65]{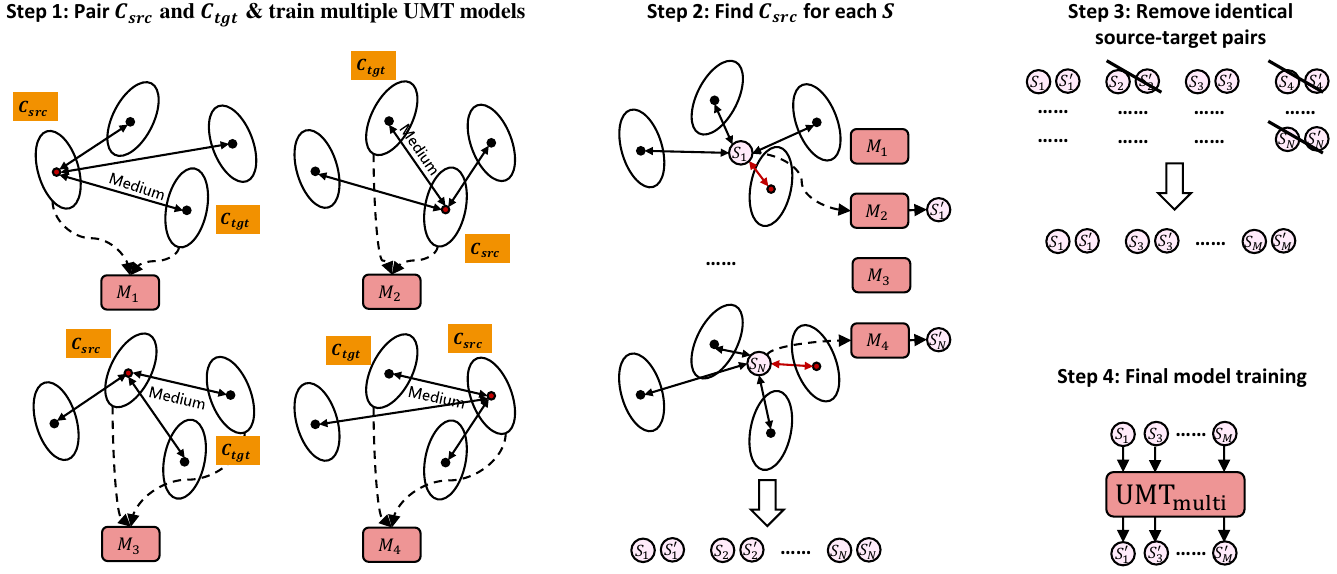}
  \caption{An overview of deriving the {\it UMT-Multi} model. Step 1: First, for each cluster $c_i$, we treat it as $C_{src}$ and find its corresponding cluster $C_{tgt}$, and then train a UMT model on ($C_{src}$, $C_{tgt}$). The number of total UMT models is $K$ (in this figure, $K=4$). Step 2: For a given input sentence $s$, we first select the $C_{src}$ that $s$ belongs to, and use the model trained on $C_{src}$ to generate the  paraphrase of $s$. This process goes on over the entire corpus, leading to a pseudo labeled dataset of paraphrase pairs. Step 3: 
  Human intervenes by 
  removing paraphrase pairs whose  inputs and output are the same, and outputs are two times longer than sources.  Step 4: Training the single {\it UMT-Multi} model using the dataset after filtering.
  Step 5 (optional):  fine-tuning the  {\it UMT-Multi} model on the supervised paraphrase dataset in the supervised setup. 
  }
  \label{fig:overview}
\end{figure*}

To paraphrase sentence $s$, we need to find its corresponding paraphrase generation model $M\in \{M_1, M_2, ..., M_K\}$, which takes $s$ as the input and outputs its paraphrase. 
We first select the $C_{src} \in \{c_1, c_2, ..., c_K\}$ that $s$ belongs to. 
Next, we pick that model $M$ trained using $C_{src}$ as sources, and use $M$ to generate the output.

For LDA, $C_{src}$ is the topic that generates $s$ with the largest probability:
\begin{equation}
C_{src} = \argmax_{c\in \{c_1, c_2, ..., c_K\}} p(s|c)
\end{equation}
For the K-means model, $C_{src}$ is the cluster whose center is closest to $s$:
\begin{equation}
C_{src} = \argmin_{c\in \{c_1, c_2, ..., c_K\}}  ||h_s - \mu_c||^2
\end{equation}
where $\mu_c$ denotes the center of the cluster $c$.

We follow \citet{lample2017unsupervised} to implement each of the UMT models.
We used the transformer \cite{vaswani2017attention} as the backbone instead of LSTMs in \newcite{bahdanau2014neural}, 
where the number of encoder blocks, 
decoder blocks, 
the number of heads,  $d_{model}$ and $d_{ff}$
are respectively set to 6, 6, 8, 512 and 2,048. 
For UMT models based on specific $C_{src}$ and $C_{tgt}$, 
both the encoder and the decoder are trained using Adam \citep{kingma2014adam}, with the learning rate set to 0.00025, $\beta_1$ set to 0.5.
We  evenly alternate between
the encoder-decoder and the discriminator.

\paragraph{Unifying $M$s into a Surrogate Model}
\label{Unifying}
 We need to  maintain $K$ different 
domain-specific 
UMT models, which is both memory costly and computationally  intensive, especially for online services. We thus propose to unify different $M$s into a single surrogate one. 
For each sentence $s$ in a selected corpus, we first find the cluster $C_{src}$ it belongs to using LDA or K-means described above, and then we use the model $M$ trained on $C_{src}$ to generate the paraphrase of $s$. In this way, we are able to collect massive amounts of pseudo-labeled paraphrase pairs by treating the original sentence $s$ as
the
 source and the produced paraphrase as the target.
We collected a total number of 25 million pairs. 
 Human interventions can happen in this stage, where we can design filtering functions to remove pairs with undesired properties.
  Here, human interventions involve  (1) 
   removing   pairs with identical source and target; 
   (2) removing targets two times longer than sources. 
16 million pairs remain after filtering. 

We train a \sts model \cite{sutskever2014sequence,vaswani2017attention}
 (referred to as {\it UMT-Multi}) on the remaining pseudo-labeled data,  which  is used as the ultimate paraphrase generation model. 
We use the  Transformer-base \citep{vaswani2017attention} as the model backbone,
where the number of encoder blocks, 
decoder blocks, 
the number of heads,  $d_{model}$ and $d_{ff}$
are respectively set to 6, 6, 8, 512 and 2,048. 
We use Adam \citep{kingma2014adam} with learning rate of 1e-4, $\beta_1$ = 0.9,
$\beta_2$ = 0.999, 
and a warmup step of 4K.
Batch size is set to 256. 
This model can be directly used in the unsupervised learning setup. 
An overview of deriving the {\it UMT-Multi} model is shown in Figure \ref{fig:overview}.
Up to now, {\it UMT-Multi} is purely based on unlabeled commoncrawl corpus.

 \subsection{Supervised  Setup}
 For the supervised setup, where we have 
 pairs of paraphrases containing
  sources from a source domain and paraphrases of sources from a target domain,
  we can fine-tune the pretrained {\it UMT-Multi} model on the supervised paraphrase pairs, where we initialize the model using the 
  {\it UMT-Multi} model, and run additional iterations on the supervised dataset. 
The fine-tuned model thus shares  
 the structure with  {\it UMT-Multi}.  
Again, we use Adam \citep{kingma2014adam} for fine-tuning, with $\beta_1=0.9$,
$\beta_2=0.98$. Batch size, learning rate 
and the number of iterations
are treated as hyper-parameters and tuned on the dev set. 
At test time, beam search \cite{sutskever2014sequence,li2016simple} is used when decoding.

An additional use of the gold labeled paraphrase datasets is to help 
 to select $C_{tgt} \in\{c_1, c_2, ..., c_K\}$ that best aligns with  $C_{src}$,
while in the unsupervised setup, we can only randomly pair $C_{src}$ and $C_{tgt}$ due to the lack of training signals for pairing. 
 In the most straightforward setup, for each $C_{src} \in\{c_1, c_2, ..., c_K\}$, we can construct $K-1$ pairs $(C_{src}, c)$
by treating 
  all $c\in\{c_1, c_2, ..., c_K\}, c\neq C_{src}$ as $C_{tgt}$.  
Next, we train $K-1$ UMT models based on the pairs, and select the model that achieves the highest evaluation score on the labeled dataset. 
This strategy leads to a total number of $K\times (K-1)$ models to be trained, which is computationally prohibitive. 
We propose a simplified learning model that  maps the distance between $ C_{src}$ and $ C_{tgt}$ as inputs to output the evaluation score (here we use iBLEU) on the labeled dataset. 
 Specifically, we randomly select $L$ pairs, where $L \ll K\times (K-1)$. We train $L$ UMT models on the selected dataset pairs. 
 Using the trained UMT models, we generate paraphrases for the labeled datasets, and obtain corresponding evaluation scores. 
Based on the distance between $ C_{src}$ and $ C_{tgt}$, and the evaluation score $S(M_{(src, tgt)})$, we train a simple polynomial
 function $F$ to learn to map the distance to the evaluation score: 
\begin{equation}
S(M_{(src, tgt)}) = F (\text{Dis}(C_{src}, C_{tgt}))
\end{equation}
The function $F$ can be then used to select $C_{tgt}$ 
with highest predicted evaluation score 
for   $C_{src}$. 
 
\section{Experiments}
\subsection{Experiment Setups}
We consider both the supervised and unsupervised setups. 
There are two differences between the supervised and unsupervised setups: 
for the supervised setup, 
(1) the training data provides guidance on pairing $C_{src}$ and $C_{tgt}$; and (2) 
the pretrained model will be used as initialization and later finetuned on the labeled dataset. 
Datasets that we use for evaluation include Quora, WikiAnswers \citep{fader-etal-2013-paraphrase}, MSCOCO  \citep{lin2014microsoft} and Twitter  \citet{liu2019unsupervised}. 

For the supervised setup, we compare our proposed model to the follow baselines: {\bf ResidualLSTM} \citep{prakash-etal-2016-neural}, {\bf VAE-SVG-eq} \citep{gupta2018deep}, {\bf Pointer} \citep{see-etal-2017-get}, {\bf Transformer} \citep{vaswani2017attention} and {\bf DNPG} \citep{li-etal-2019-decomposable}.
For the unsupervised setup, we use the following models for comparison:  {\bf VAE} \citep{bowman-etal-2016-generating}, {\bf Lag VAE} \citep{he2019lagging}, {\bf CGMH} \citep{miao2019cgmh} and {\bf UPSA} \citep{liu2019unsupervised}.
Results for VAE, Lag VAE, CGMH and UPSA on different datasets are copied from \newcite{miao2019cgmh} and \newcite{liu2019unsupervised}. 
Results for ResidualLSTM, VAE-SVG-eq, Pointer, Transformer on various datasets are copied from \newcite{li-etal-2019-decomposable}. 
We leave details of these datasets, baselines and  training in Appendix \ref{appendix}.

We are particularly interested in comparing the proposed model with bilingual MT based models. 
 BT is trained end-to-end on WMT'14 En$\leftrightarrow$Fr.\footnote{\newcite{wieting2017learning,wieting2017paranmt} suggested little difference among Czech,
German, and French as source languages for back-translation. We use En$\leftrightarrow$Fr since it contains more parallel data than other language pairs. } 
A paraphrase pair is obtained by pairing the English sentence in the original dataset and the translation of the French sentence. 
Next we train a Transformer-large model on paraphrase pairs. 
The  model is used as initialization to be further finetuned on the labeled dataset.
We also use WMT-14 En-Zh for reference purposes. 
We use  BLEU \citep{papineni2002bleu}, iBLEU \citep{sun2012joint} and ROUGE scores \citep{lin-2004-rouge} for evaluation. 

\subsection{Results}
\paragraph{In-domain Results}
We first show the in-domain results in Table \ref{tab:in-domain}. We can observe that across all datasets and under both the supervised and unsupervised setups, the proposed UMT model significantly outperforms  than baselines. As expected, multiple UMT models perform better than a single UMT model as the former is more flexible at selecting the correct domain $C_{src}$ for an input sentence. We can also observe that the BT model is able to achieve competitive results, which shows that back-translation serves as a strong and simple baseline for paraphrase generation. The BT model trained on En-Fr consistently outperforms the one trained on En-Zh, and this is because that
En-Zh translation  is a harder task than En-Fr due to the greater grammars difference between the two languages. 

\begin{table}[!t]
  \centering
  \small
  \scalebox{0.95}{
  \begin{tabular}{clcccc}\toprule
    & {\bf Model}  & {\bf iBLEU} & {\bf BLEU} & {\bf R1} & {\bf R2}\\\midrule
    \multirow{22}{*}{\rotatebox{90}{{Supervised}}} & \multicolumn{5}{c}{\underline{\it Quora}}\vspace{1pt}\\
    & {\it ResidualLSTM} & 12.67  & 17.57 &  59.22 & 32.40\\
    & {\it VAE-SVG-eq} &15.17 & 20.04 & 59.98 & 33.30\\
    & {\it Pointer} &  16.79 & 22.65 & 61.96 & 36.07\\
    & {\it Transformer} & 16.25 & 21.73 & 60.25 & 33.45\\
    & {\it DNPG} & 18.01 & 25.03 & 63.73 & 37.75\\
    
    & {\it BT}(En-Fr) &  18.04 & 25.34 & 63.82 & 37.92 \\
    & {\it BT}(En-Zh) & 17.67&24.90& 63.32& 37.38 \\
    &{\it UMT-Single}&  17.70 & 24.97& 63.65 & 37.77  \\
    & {\it UMT-Multi} & \bf{18.78}& \bf{26.49} & \bf{64.12} & \bf{38.31} \\

    \specialrule{0em}{1pt}{1pt}
    \cdashline{2-6}
    \specialrule{0em}{1pt}{1pt}
    & \multicolumn{5}{c}{\underline{\it Wikianswers}}\vspace{1pt}\\
    & {\it ResidualLSTM} & 22.94& 27.36 &48.52& 18.71    \\
    & {\it VAE-SVG-eq} &26.35 &32.98 &50.93 &19.11\\
    & {\it Pointer} & 31.98 &39.36 &57.19 &25.38\\
    & {\it Transformer} & 27.70& 33.01& 51.85& 20.70\\
    & {\it DNPG} & 34.15& 41.64& 57.32& 25.88\\
    
        & {\it BT}(En-Fr) &  34.55 & 41.90 & 57.84 & 26.44 \\
    & {\it BT}(En-Zh) &33.98 & 41.04 & 56.37 & 25.60  \\
    &{\it UMT-Single} &  34.50 & 41.72 & 57.58 & 26.31  \\
    & {\it UMT-Multi} & \bf{36.04}& \bf{42.94} & \bf{58.71} & \bf{27.35} \\
    \cmidrule{1-6} 
    \multirow{36}{*}{\rotatebox{90}{{Unsupervised}}} & \multicolumn{5}{c}{\underline{\it Quora}}\vspace{1pt}\\
    & {\it VAE} & 8.16& 13.96 &44.55 &22.64 \\
    & {\it Lag VAE} & 8.73& 15.52 &49.20 &26.07 \\
    & {\it CGMH} & 9.94& 15.73& 48.73 &26.12 \\
    & {\it UPSA} & 12.03 &18.21 &59.51 &32.63  \\
    & {\it BT}(En-Fr) & 11.98& 17.84 & 59.03 & 32.11 \\
    & {\it BT}(En-Zh) & 11.33& 17.02 & 56.19 & 31.08 \\
    &{\it UMT-Single} &  11.47 &17.21 & 56.35 &31.27  \\
    & {\it UMT-Multi} & \bf{13.10}& \bf{18.98} & \bf{59.90} & \bf{33.04} \\
        \specialrule{0em}{1pt}{1pt}
    \cdashline{2-6}
    \specialrule{0em}{1pt}{1pt}
    & \multicolumn{5}{c}{\underline{\it Wikianswers}}\vspace{1pt}\\
    & {\it VAE} &  17.92 &24.13 &31.87 &12.08 \\
    & {\it Lag VAE} & 18.38& 25.08 &35.65 &13.21 \\
    & {\it CGMH} &  20.05 &26.45 &43.31 &16.53 \\
    & {\it UPSA} & 24.84& 32.39 &54.12 &21.45\\
    & {\it BT}(En-Fr) & 23.55& 31.10 & 52.03 & 20.86 \\
    & {\it BT}(En-Zh) & 22.60& 30.12 & 51.29 & 20.11 \\
    &{\it UMT-Single} &  23.01 &30.62 & 51.79 &20.35  \\
    & {\it UMT-Multi} & \bf{25.90}& \bf{33.80} & \bf{54.52} & \bf{23.48} \\
    \specialrule{0em}{1pt}{1pt}
    \cdashline{2-6}
    \specialrule{0em}{1pt}{1pt}
    & \multicolumn{5}{c}{\underline{\it MSCOCO}}\vspace{1pt}\\
    & {\it VAE} & 7.48 &11.09& 31.78& 8.66 \\
    & {\it Lag VAE} &  7.69 &11.63 &32.20 &8.71 \\
    & {\it CGMH} & 7.84& 11.45& 32.19 &8.67 \\
    & {\it UPSA} & 9.26 &14.16 &37.18& 11.21 \\
        & {\it BT}(En-Fr) & 8.15& 13.78 & 36.30 & 10.48 \\
    & {\it BT}(En-Zh) & 7.80&11.97 & 32.40 & 9.21 \\
    &{\it UMT-Single} &  8.21 &13.99 & 36.52 &10.75  \\
    & {\it UMT-Multi} & \bf{9.70}& \bf{15.42} & \bf{38.51} & \bf{12.39} \\
    \specialrule{0em}{1pt}{1pt}
    \cdashline{2-6}
    \specialrule{0em}{1pt}{1pt}
    & \multicolumn{5}{c}{\underline{\it Twitter}}\vspace{1pt}\\
    & {\it VAE} & 2.92& 3.46& 15.13& 3.40    \\
    & {\it Lag VAE} & 3.15& 3.74& 17.20 &3.79 \\
    & {\it CGMH} & 4.18& 5.32& 19.96 &5.44\\
    & {\it UPSA} &  4.93& 6.87 &28.34 &8.53 \\
    
        & {\it BT}(En-Fr) & 4.32& 5.97 & 26.37 & 7.59 \\
    & {\it BT}(En-Zh) & 4.15&5.40 & 25.83 & 7.32 \\
    &{\it UMT-Single} &  4.40 &6.11 & 26.89 &7.78  \\
    & {\it UMT-Multi} & \bf{5.35}& \bf{7.80} & \bf{29.74} & \bf{9.88} \\
    
    \bottomrule
  \end{tabular}
  }
  \caption{In-domain performances of different models for both supervised and unsupervised setups. 
  }
  \label{tab:in-domain}
\end{table}

\begin{table}[!t]
  \centering
  \small
  \scalebox{0.95}{
  \begin{tabular}{lcccc}\toprule
      {\bf Model} & {\bf iBLEU} & {\bf BLEU} & {\bf R1} & {\bf R2}\\\midrule 
      \multicolumn{5}{c}{\underline{\it Wikianswers$\rightarrow$Quora}}\\
        {\it Pointer}& 5.04 &6.96& 41.89 &12.77 \\
        {\it Transformer+Copy} &  6.17& 8.15& 44.89 &14.79 \\
        {\it DNPG} &10.39 &16.98 &56.01& 28.61\\
               {\it BT}(En-Fr) &12.14& 17.98 & 59.42 & 32.44 \\
     {\it BT}(En-Zh) & 11.43& 17.21 & 56.65 & 31.45 \\
    {\it UMT-Single} &  11.86 &17.49 & 57.01 &32.44  \\
    {\it UMT-Multi} & \bf{13.62}& \bf{19.48} & \bf{61.04} & \bf{33.85} \\
        \specialrule{0em}{1pt}{1pt}
        \cdashline{1-5}
        \specialrule{0em}{1pt}{1pt}
    \multicolumn{5}{c}{\underline{\it Quora$\rightarrow$Wikianswers}}\\
        {\it Pointer}  &21.87 &27.94& 53.99 &20.85 \\
        {\it Transformer+Copy}&  23.25& 29.22& 53.33& 21.02 \\
        {\it DNPG}  & 25.60 & 35.12 &56.17 &23.65 \\
        
       {\it BT}(En-Fr) &25.77& 35.30 & 56.41 & 23.78 \\
     {\it BT}(En-Zh) & 24.84& 34.19 & 55.71 & 22.60 \\
    {\it UMT-Single} &  25.43 &34.70 & 56.10 &23.31  \\
    {\it UMT-Multi} & \bf{26.85}& \bf{36.64} & \bf{57.45} & \bf{24.60} \\\bottomrule

  \end{tabular}
  }
  \caption{Domain-adapted performances of different models.
   ``R1'' stands for ROUGE-1 and ``R2'' stands for ROUGE-2.}
  \label{tab:cross-domain}
\end{table}

\begin{table*}[!t]
  \centering
  \small
  \scalebox{0.9}{
  \begin{tabular}{p{5.5cm}p{5.5cm}p{5.5cm}}
  \toprule
  {\bf Input} & {\bf BT} & {\bf UMT} \\
   \midrule
   Time is the most accurate test of how puberty is going to progress. & Time is the most accurate test of how puberty is going to progress.  (En-Fr)  & 
   How adolescence develops is most accurately test by time. 
    \\\cline{1-3}
   GM told truck owners about the defect in the first half of October. & GM owners told truck about the defect in the first half of October . (En-Fr)  &GM informed truck owners of the defect in the first half of October . \\\cline{1-3}
   To keep him alive , well , there 's no reason to keep him alive . &  To keep him alive , well , there 's no reason to keep him alive . (En-Fr) &  To let him live , well , there 's no reason to let him live .   \\\cline{1-3}
      Washington did not have the formal education his elder brothers received at Appleby Grammar School in England, but he did learn mathematics, trigonometry, and land surveying .  &
   Washington did not pursue the same formal education as his older brothers at England's Appleby Grammar School, but he did study geometry, trigonometry, and land surveying .&
   Unlike his older brothers, who studied at England's Appleby Grammar School, Washington
   did not receive 
   formal education but
    studied mathematics, trigonometry, and land surveying .\\ 
   \bottomrule
  \end{tabular}
  }
  \caption{Sampled paraphrases from the BT and  UMT models.}
  \label{tab:example}
\end{table*}

\paragraph{Domain-adapted Results}
We test the model's domain adaptation ability on Quora and Wikianswers. Table \ref{tab:cross-domain} shows the results.
We can see that UMT-multi performs significantly better than baselines, including UMT-single, showing the better ability of UMT-multi for domain adaptation.

\subsection{Human Evaluation}
To further validate the performance of the proposed model, we randomly sample 500  
sentences from Quora test set for human evaluation.
The input sentence and its two paraphrases respectively generated by the UMT model and the BT model (En-Fr)
 are  assigned to two human annotators at Amazon Mechanical Turk (AMT), with “> 95$\%$ HIT approval rate”.
Annotators are asked to judge which output is better in terms of three aspects:
(1) semantics: whether the two sentences are the same semantic meaning;
(2) diversity: whether the two sentences are diverse in expressions; and 
(3) fluency: whether the generated paraphrase is fluent. 
Ties are allowed. 
If the two annotators' evaluations  do not agree with each other, 
the job will be assigned to one more annotator, and we take the majority  as the final result.\footnote{
If the three annotators all disagree, we discard the instance.} 
Comparing with BT, the proportions of win, tie and lose for the proposed  UMT-model are respectively 
$41\%$, $36\%$, and $22\%$, demonstrating its superiority over BT  models. 

\subsection{Examples}

Table \ref{tab:example} presents sampled paraphrases from the BT and  UMT  models. From these examples, we can identify several intrinsic drawbacks of the BT model  that  the UMT model can circumvent: 
(1) for the first example, the tense from the BT paraphrase model based on En-Zh translation is incorrect. This is because the Chinese language expresses tense in a more implicit way. 
This leads the model to make mistake in tense 
when Chinese is translated back to English. The UMT model does not have this issue; 
(2) for the second example, BT model directly copies the input, this is because the En-Fr can perfectly map the meaning in two languages with no expression variations. 
Due to the blackbox nature of MT models, it is hard to intervene with the process to avoid producing the same copy. 
Instead, for the proposed UMT framework, developers can intervene with the model in both clustering stage and data filtering stage. 
(3) For the third example, the BT model changes the meaning of the original sentence, which is due to the mistake made by the translation model. 
These mistakes are sometimes inevitable due to the limitation of current MT models, but can be fixed in the proposed system.

\section{Ablation Study}
In this section, we perform comprehensive ablation studies on Wikianswers dataset for understanding  behaviors of the proposed model.  
And we report iBLEU score for comparison.
\paragraph{Size of $C$ for UMT Training}
First, we explore how the size of $C$, the CommonCrawl corpus used for dataset split and UMT training, affects downstream performances. 
Table \ref{tab:size} shows the results, where the size is respectively 10M, 100M, 1B and 10B.
We can
observe that with more training data, 
the performance significantly improves.
This is
because  
the trained model can better learn to align sentences between different clusters.

\begin{table}[t]
    \centering
    \small
    \scalebox{0.9}{
    \begin{tabular}{lccccc}\toprule
    {\bf Size} & {\bf 10M} & {\bf 100M} & {\bf 1B} & {\bf 10B} \\\midrule 
    {\it Unsupervised. UMT-Multi} & 15.5 & 21.1 & 24.2 & {\bf 25.9}  \\
    \bottomrule
    \end{tabular}
    }
    \caption{The effect of data size of $C$ for training UMT. }
    \label{tab:size}
\end{table}

\paragraph{The Number of LDA Topics}
Table \ref{tab:number} presents the influence of the number of LDA clusters. The trend is clear: more topics lead to better performances. 
This is because the model with more topics has a stronger ability of disentangling very similar sentences in the original corpus $C$, and thus avoids copying. 
It is worth noting that more topics means training more UMT models before unifying them, leading to greater computational intensity. 

\begin{table}[t]
  \centering
  \small
  \scalebox{0.9}{
  \begin{tabular}{lccccc}\toprule
  {\bf \# LDA Topic} & {\bf 5} & {\bf 20} & {\bf 50} & {\bf 80} \\\midrule 
  {\it Unsupervised. UMT-Multi} & 14.9 & 22.4 & 24.9 & {\bf 25.9}  \\
  \bottomrule
  \end{tabular}
  }
  \caption{The effect of number of LDA topics. }
  \label{tab:number}
\end{table}

\paragraph{Pairing $C_{src}$ and $C_{tgt}$}
In our main experiments, we 
randomly select
 $C_{tgt}$ given $C_{src}$.
It would be interesting to see the effects of different cluster selection strategies. We consider four strategies:
 {\it Largest} (select  $C_{tgt}$ with the largest distance to $C_{src}$), {\it Medium} (select  $C_{tgt}$ with the medium distance to $C_{src}$), {\it Smallest} (select  $C_{tgt}$ with the smallest distance to $C_{src}$) and 
for referring purposes, 
{\it Supervised} (select  $C_{tgt}$ using the supervised strategy proposed). In the {\it real} unsupervised setup, the supervised strategy cannot be readily applied since we have no access to supervised labels. We list performance for {\it supervised} here for referring purpose.
 
Table \ref{tab:pair} shows the results. For both supervised and unsupervised setups, {\it Supervised} performs the best against the other  strategies, especially under the unsupervised setup.
The difference in performances  between these strategies is greater for the unsupervised setup than the supervised setup. 
This is because supervised training serves to compensate the performance gap due to the presence of labeled training data.
We find that the random strategy outperforms both   {\it Largest} and   {\it Smallest}.
For {\it Largest} ,
this is because  {\it Largest}  leads to very different paired clusters, having the risk that some sentences in $C_{src}$ might not have correspondences in 
$C_{tgt}$. For
{\it Smallest}, since paired clusters are pretty close, sentences in $C_{src}$ are more likely to have copies in $C_{tgt}$.
    {\it Largest} and   {\it Smallest} leads to inferior performances.
{\it random} performs comparable to {\it medium}. 
\begin{table}[t]
  \centering
  \small
  \scalebox{0.9}{
  \begin{tabular}{lcc}\toprule
  {\bf Strategy}  & {\bf Unsuper. UMT-Multi} & {\bf Super. UMT-Multi}\\\midrule
  {\it Random} & 25.9 & 35.4\\
  {\it Largest} & 24.7 & 35.1\\
  {\it Medium} & 25.8 & 35.7\\
  {\it Smallest} & 25.3 & 35.5\\
  {\it Supervised} & {\bf 26.3} & {\bf 36.0}\\
  \bottomrule
  \end{tabular}
  }
  \caption{The effect of different strategies to pair $C_{src}$ and $C_{tgt}$. }
  \label{tab:pair}
\end{table}

 \paragraph{Clustering Methods}
We  study the effect of different clustering methods, i.e., LDA and K-means.
Table \ref{tab:cluster} shows the results. As can be seen, for both supervised and unsupervised setups, the model trained with LDA consistently performs better than the model trained with K-means.
We think there are potentially two reasons: (1) the BERT representations, on which clustering relies, cannot well represent sentence semantics for clustering \citep{reimers2019sentence}; and (2) the K-means model for sentence clustering operates at a relatively low level of semantics (i.e., sentence level), while LDA takes into the more global document level information. 
Due to the  entanglement of sentence semantics in $C$, it is  hard for K-means to separate sentences apart, or if it can, it  takes   long until convergence. 

\begin{table}[t]
  \centering
  \small
  \scalebox{0.9}{
  \begin{tabular}{lcccc}\toprule
    & \multicolumn{2}{c}{{\bf LDA}} & \multicolumn{2}{c}{{\bf K-means}} \\
  {\bf Clustering} & {\bf Single} & {\bf Multi} & {\bf Single} & {\bf Multi} \\\midrule 
  {\it Uns.} & 23.0& {\bf 25.9} & 21.9 & {24.2}  \\
  {\it Su.} & 34.5 & {\bf 36.0} & 32.1 & {34.2}  \\
  \bottomrule
  \end{tabular}
  }
  \caption{The effect of different clustering methods for $C$. 
  ``Uns.'' means we use the unsupervised setup and ``Su.'' represents the supervised setup.}
  \label{tab:cluster}
\end{table}

\section{Conclusion}
In this paper, we propose a new framework for paraphrase generation by treating the task as unsupervised machine translation (UMT). The proposed framework first splits a large unlabeled corpus into multiple sub-datasets and then trains one or multiple UMT models based on one or more pairs of these sub-datasets.
 Experiments and ablation studies under  supervised and unsupervised setups demonstrate the effectiveness of the proposed framework.
 
\section*{Acknowledgement}
We would like to thank anonymous reviewers for their comments and suggestions.
This work is supported by the Key R \& D Projects of the Ministry of Science and Technology (2020YFC0832500).

\bibliography{custom}
\bibliographystyle{acl_natbib}

\newpage

\appendix 
\label{appendix}
\section{Datasets}
(1) {\bf Quora}\footnote{\url{https://www.kaggle.com/c/quora-question-pairs}}: The Quora dataset contains 140K parallel paraphrases of questions and 260K non-parallel sentences collected from the  question answering website Quora\footnote{\url{https://www.quora.com/}}.  We follow the standard setup in \citet{miao2019cgmh} and  use 3K/30K paraphrase pairs respectively for validation and test. \\
(2) {\bf Wikianswers}: The WikiAnswers corpus \citep{fader-etal-2013-paraphrase} contains clusters of questions tagged by WikiAnswers users as paraphrases. It contains a total number of 2.3M paraphrase pairs. We follow \citet{liu2019unsupervised} to randomly pick 5K pairs for validation and 20K for test.\footnote{Note that the selected data is different from \citet{liu2019unsupervised} but is comparable in the statistical sense.} \\
(3) {\bf MSCOCO}: The MSCOCO dataset \citep{lin2014microsoft} contains over 500K paraphrase pairs for 120K image captions, with each image caption annotated by five annotators. We follow the  dataset split  and the evaluation protocol in \citet{prakash-etal-2016-neural}, where only image captions with fewer than 15 words are considered.\\
(4) {\bf Twitter}: The Twitter dataset is collected via linked tweets through shared URLs \citep{lan-etal-2017-continuously}, which originally contains 50K paraphrase pairs. We follow the data split in \citet{liu2019unsupervised}, where 10\% of the training data is used as validation and the test set only contains sentence pairs that are labeled as  ``paraphrases''. 

\section{Baselines}

For the supervised setup, we compare our proposed model to the follow baselines:\\
(1) {\bf ResidualLSTM}: \citet{prakash-etal-2016-neural} deepened the LSTM network by stacking multiple layers with residual connection. This deep \sts model is trained on labeled paraphrase datasets.\\
(2) {\bf VAE-SVG-eq}: \citet{gupta2018deep} combined VAEs with LSTMs to generate paraphrases in a \sts generative style. \\
(3) {\bf Pointer}: \citet{see-etal-2017-get} augmented the standard \sts model by using a pointer, i.e., the copy mechanism. Word in the input sentence can be directly copied as the current decoded word.\\ 
(4) {\bf Transformer}: \citet{vaswani2017attention} proposed the Transformer architecture which is based on the self-attention mechanism. \\
(5) {\bf DNPG}: \citet{li-etal-2019-decomposable} proposed a Transformer-based model that learns and generates paraphrases at different levels of granularity, i.e., from the lexical to phrasal and then to sentential levels.

For the unsupervised setup, we use the following models for comparison:\\
(1) {\bf VAE}:  \citet{bowman-etal-2016-generating} proposed variational auto-encoders (VAEs) to generate sentences from a continuous space. By minimizing the reconstruction loss between the input sentence and the output sentence, VAEs are able to sample paraphrases from the continuous space.\\
(2) {\bf Lag VAE}: To overcome the posterior collapse issue of VAEs, \citet{he2019lagging} proposed to aggressively optimize the inference network by performing multiple updates before reverting back to basic VAE training. 
\\
(3) {\bf CGMH}: \citet{miao2019cgmh} used Metropolis–Hastings sampling to generate paraphrases, where a word can be deleted, replaced or inserted into the current sentence based on the sampling distribution. 
\\
(4) {\bf UPSA}: \citet{liu2019unsupervised} proposed to use simulated annealing to optimize the  paraphrase generation model. The training objective is composed of three parts: semantic similarity, expression diversity and language fluency.

\end{document}